\documentclass[preprintnumbers,amsmath,amssymb,prl,final]{revtex4}
\usepackage{bm}% bold math

\usepackage{amsmath}
\usepackage{amssymb}
\usepackage{amsfonts}   
\usepackage{graphics}  
\usepackage{graphicx} 
\usepackage{epsfig}    
 \usepackage[latin1]{inputenc}
\usepackage[usenames,dvipsnames]{color}
 \usepackage{rotating}
 \usepackage{wrapfig}

\begin{document}
\nocite{*}
%\preprint{}

\title{Infomax strategies for an optimal balance between exploration and exploitation}
\author{Gautam Reddy}
\affiliation{University of California San Diego, Department of Physics, La Jolla, CA 92093 USA}
\author{Antonio Celani}
\affiliation{The Abdus Salam International Centre for Theoretical Physics (ICTP), Strada Costiera 11, I-34014 - Trieste, Italy}
\author{Massimo Vergassola}
\affiliation{University of California San Diego, Department of Physics, La Jolla, CA 92093 USA}

\date{\today}

\begin{abstract}
Proper balance between exploitation and exploration is what makes good decisions, which achieve high rewards like payoff or evolutionary fitness. The Infomax principle postulates that maximization of information directs the function of diverse systems, from living systems to artificial neural networks. While specific applications are successful, the validity of information as a proxy for reward remains unclear. Here, we consider the multi-armed bandit decision problem, which features arms (slot-machines) of unknown probabilities of success and a player trying to maximize cumulative payoff by choosing the sequence of arms to play. We show that an Infomax strategy (Info-p) which optimally gathers information on the highest mean reward among the arms, saturates known optimal bounds and compares favorably to existing policies. The highest mean reward considered by Info-p is not the quantity actually needed for the choice of the arm to play, yet it allows for optimal tradeoffs between exploration and exploitation. 
\end{abstract}

%\keywords{}

\maketitle

\section{Introduction}
Shannon's theory of information deliberately leaves aside the meaning of 
messages and focuses on their statistical properties \cite{Shannon48}. 
%The introduction to his seminal 
%work 
%reads indeed\,: ``These semantic aspects of communication are irrelevant to the engineering problem. The significant aspect 
%is that the actual message is one selected from a set of possible messages.'' 
This standpoint is crucial for the universality of the theory, as witnessed by its wide 
range of applications in communication, computation and learning \cite{Gallager,Mezard,MacKay}. 

Biological and economic sciences feature 
natural measures of ``meaning'', i.e. evolutionary fitness and payoffs. The relation between payoffs and information was first addressed by Kelly for his model of horse race gambling \cite{Kelly}, where information on the outcome 
of the race provides
 a bound on the increment in the doubling rate of returns. 
 %and this difference is always positive, i.e. 
% information can be irrelevant but it never hurts. 
The question was further developed and applied to portfolio management in Refs.~\cite{Howard,Barron,Cover}. Kelly's horse race appears in some aspects of evolutionary biology as well. There, the reward function is the population growth rate and information 
refers to the state of the environment \cite{Bergstrom,Kussell,Donaldson-Matasci,Rivoire,Bialek}. 
%In this framework, the mutual information between environmental states and cues sets an upper limit to the difference in the maximal growth-rates that can be achieved by a population that detects cues with respect to one that does not 

Neurobiology is the field where information theory is arguably the most popular in biological sciences. Barlow's efficient coding  \cite{Barlow61} postulated that 
early neural sensory layers efficiently represent environmental information, i.e. their evolutionary fitness is proportional to 
their efficiency in the transmission of information from the environment to higher parts of the brain.  The hypothesis was 
spectacularly confirmed in the visual system \cite{Laughlin89,Atick92,Rieke}, see also \cite{Bialek,Dayan}. Similar ideas were recently introduced 
in cellular biology, namely to transduction pathways \cite{Cheong}, their computational inference \cite{Margolin} and evolution \cite{Francois}, adaptation \cite{Nemenman,Sharpee} and transcription regulation \cite{Tkacik,Walczak}. 

The catchy name Infomax for the maximization of information was introduced in \cite{Linsker88}, where it was 
applied to the training of perceptual networks. Infomax was also later applied for blind separation and deconvolution \cite{Bell95}. Infotaxis \cite{infotaxis} used information as an orientation cue for searches aimed at locating sources of chemicals transported in a turbulent environment. For a recent review of information theory for decisions and actions, see \cite{Tishby}. 

%Works above have led to the sense that information theory is relevant for living systems and their decisions. 
It is usually the case that the more information is available, the better decisions or performances are, e.g. for the evolutionary model  discussed in \cite{Bialek} the fitness increases with available information on the state of the environment. However, acquiring information has costs so that maximizing information does not generally lead to the best decisions. A first reason is the direct cost of acquiring and processing information, e.g. energy consumption costs\,:
random strategies can obviously be the most effective if those costs are too high. 
The second, more subtle cost is that the choice of an action entails the exclusion of other possibilities. That calls for a balance between exploration and exploitation \cite{Sutton}, which is what we shall discuss in the sequel. 

Decisions in fluctuating and unknown environments require a balance between two extremes\,: exhaustive exploration of all available options {\it vs} greedy exploitation of available information to maximize short-term return. 
While the first option seems wiser, it can still performs poorly as harvesting information does not coincide with maximizing reward. For instance, for the search problem of a source of chemicals discussed in \cite{infotaxis}, the actual quantity to be minimized is the time of completion of the search. The information on the location of the source was found to be an efficient proxy, which replaces the daunting estimation of completion times by a much simpler statistic. When is such a replacement possible? More generally, when is the Infomax principle applicable and what are the situations, if any, where it is optimal? 

Here, we address the previous questions by considering a classical problem in statistical decision 
theory: the multi-armed bandits. The model is the prototype of a broad class of sequential allocation problems that aim at optimally dividing resources to projects which yield benefits at a rate that depends on their degree of development. Among its many applications, we mention clinical trials, adaptive routing, job-scheduling, portfolio design and military logistics (see \cite{berry,gittins} and references therein). Beside practical applications, the multi-armed bandit problem embodies the dilemma between exploitation and exploration mentioned above~\cite{whittle}. The additional appeal of the model
is that optimal strategies of decision are known, the so-called Gittins index \cite{Gittins_paper}, as well as asymptotic bounds on maximal gains \cite{lairob}. That allows to gauge the performance of the Infomax strategies developed below, Info-p and Info-id, and to provide a systematic assessment of cost and value of information. Finally, optimal bounds for the multi-armed bandit can be generalized to the broader class of problems encompassing Markov Decision Processes \cite{Burnetas97}, suggesting that methods developed for the multi-armed bandit problem can have general relevance. To facilitate reading, we shall first briefly review known relevant results and then present our  own.

%We first briefly introduce the multi-armed bandit problem and the main results that will be relevant here, and then present the novel Infomax strategies and the corresponding results.
\section{The multi-armed bandit problem in a nutshell} At each discrete time, an agent chooses to pull one arm among $K$ available. The agent receives a reward for the chosen action, according either to some unknown distribution or to a known distribution  with unknown parameters. 
We shall consider for concreteness the case of $K$ Bernoulli arms whose (unknown) probabilities of success are $p_1, p_2,\dots ,p_K$, which are ordered for future convenience as $p_1 > p_2>\dots >p_K$. After each play, a reward is paid, which is (rescaled to) unity upon winning and zero otherwise. The long-term goal is to find a strategy that maximizes the average cumulated reward or, equivalently, minimize the expected regret $R$\,:
%\begin{align}
$R(p_1,p_2,\dots,p_k) = \sum  \overline{n}_i (p_1 - p_i)$,
%\end{align}
where $\overline{n}_i$ is the expected number of plays of the $i$th arm. 

Gittins index policy \cite{Gittins_paper} applies to discounted rewards, i.e. maximizes the expected value of the sum $r_0 + \gamma r_1  + \gamma^2 r_2 + \dots$ where $\gamma$ is a discount factor between zero and one. Even though the total number of steps is infinite, the discount factor introduces an effective horizon $\propto (1 - \gamma)^{-1}$.  For this formulation, Gittins \cite{Gittins_paper} showed that the optimal strategy is an index policy, i.e. for each arm $i$, one computes an index independent of all other arms, and then plays the arm with the highest index. The expression of the Gittins 
index $\nu_i$ for the $i$-th arm at time $t$ is
\begin{equation}
\nu_{i}(w_i,n_i)={\rm sup}_{\tau>0}\frac{\langle\sum_{k=0}^{\tau-1} \gamma^k r_{t+k+1}\rangle}
{\langle\sum_{k=0}^{\tau-1} \gamma^k\rangle}\,,
\label{eq:Gittins}
\end{equation}
where $r_{t+1+k}$ are the future rewards that one would obtain by choosing to play uniquely the $i$-th arm up to the stopping time $t+\tau$. 
The brackets in \eqref{eq:Gittins} denote the expectations of future success based on the posterior distribution defined by the past outcomes $w_i$ and $n_i$ (see \eqref{eq:posterior_beta}).  Finally, the sup in \eqref{eq:Gittins} is taken over future stopping times, i.e. decisions that interrupt the game based only on information obtained up to the stopping time. In other words, the Gittins index \eqref{eq:Gittins} yields the expected rate of future rewards for the $i$-th arm, given 
its past number of plays $n_i$ and wins $w_i$.
While \eqref{eq:Gittins} is the only expression consistent with an index policy (see Chap. 2 in \cite{gittins}), the existence of an index policy itself is remarkable, and it is specific to the discounted formulation. The calculation of the Gittins index is usually done via dynamic programming \cite{gittins}. However, the exponentially growing 
number of possible paths makes the problem intractable as the discount factor $\gamma$ approaches unity.

The Lai-Robbins \cite{lairob} lower bound on the expected number of plays of suboptimal arms reads\,: 
\begin{equation}\label{eq:lairob}
\overline{n}_i \ge \frac{\ln n }{D(p_i, p_1)} + \text{terms of lower order in}\,\, n\,.
\end{equation}
The bound is generally valid when the number of plays $n$ is large and it does not involve any discount. In \eqref{eq:lairob}, $i \ne 1$ and $D(p, q)$ is the Kullback-Leibler relative entropy, that is the standard measure of divergence between two probability 
distributions \cite{Cover}. Specifically, $D(p,q) = p\ln \frac{p}{q} + (1-p)\ln \frac{1-p}{1-q}$ for two Bernoulli distributions parameterized by $p$ and $q$. 
The closer the two probabilities $p_1$ and $p_i$ are, the larger is the constant in \eqref{eq:lairob} and $ \overline{n}_i \propto \ln n/\left(p_1-p_i\right)^2$ as $p_i\to p_1$. Strategies that attain the bound \eqref{eq:lairob} are called asymptotically optimal.

\section{Results}

Hereafter, we introduce two Infomax strategies, Info-p and Info-id. 
%, analyze their performance and exploit them to quantify cost and value of information. 
Info-p greedily acquires information on the estimated highest success probability among the arms of the bandit. We show below that Info-p saturates the bound \eqref{eq:lairob}, i.e. it is asymptotically optimal. Conversely, Info-id gathers information about the identity of the best arm. While Info-p leads to optimal payoffs,  Info-id is shown below to yield an optimal rate of acquisition of information on the identity of the best arm but suboptimal payoffs.

\subsection{Info-p}
\begin{figure}
\begin{center}	
\includegraphics[width=.4\textwidth]{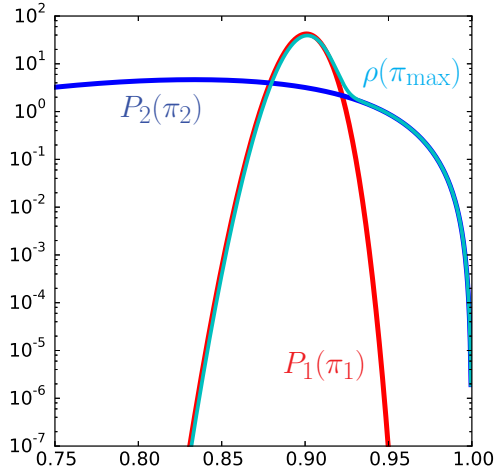}
\caption{The posterior distributions $P$ for the estimated probabilities of success $\pi_1$, $\pi_2$ of a two-armed bandit and the corresponding distribution of $\pi_{\max}={\max}_i\pi_i$. The total number of plays is $n = 1000$ and the number of plays of the suboptimal arm $n_2 = 20$. By the law of large numbers, the two distributions (in red for $\pi_1$ and blue for $\pi_2$) 
are typically centered around their respective true values of the probabilities $p_1=0.9$ and $p_2=0.8$. 
The core of the two distributions is Gaussian by the central limit theorem, while far tails are controlled by large deviations theory \cite{Cover}. 
The resulting distribution of $\pi_{\max}$ (in cyan) has a Gaussian bulk and a right tail that is controlled by the suboptimal arm. The tail captures the probability of misclassifying the order of the arms, as explained in the text.}\label{fig:infop}
\end{center}
\end{figure}

Unless specified otherwise, we discuss for simplicity a two-armed bandit with success probabilities $p_1>p_2$. Results are easily 
generalized to $K$ arms. The probability of success for the $i$th arm, as estimated from a sample of plays,
is denoted by $\pi_i$. 
%in this case, the model is analytically solvable, and the algorithms and analysis can be extended to the case of $k$ bandits. 
Its posterior distribution $P_i(\pi_i)$ after $n_i$ plays and $w_i$ wins reads (see, e.g., \cite{MacKay})\,:
\begin{equation}
P_i(\pi_i) = \frac{\pi_i^{w_i} (1-\pi_i)^{n_i-w_i}}{B(w_i + 1, n_i- w_i + 1)}\,,
\label{eq:posterior_beta}
\end{equation}
where $B$ is the Euler $\beta$-function. In \eqref{eq:posterior_beta} we assumed a uniform prior\,; a different prior requires minor modifications and does not affect subsequent results.  
We are interested in the distribution of $\pi_{\max} = \max_i \pi_i$, i.e. the largest success probability among the arms of the bandit. The probability density $\rho(\pi_{\max})$ is the sum of the contributions by each arm, weighted by the probability for that arm to be the best:
\begin{equation}
\label{eq:rho}
\rho(\pi_{\max}) = P_1(\pi_{\max}) \int_0^{\pi_{\max}} \!\!\!\!\!\!\!\!\!\!\!P_2(p) dp + P_2(\pi_{\max}) \int_0^{\pi_{\max}} \!\!\!\!\!\!\!\!\!\!\!P_1(p) dp\,.
\end{equation}
Fig.~\ref{fig:infop} shows the posterior distributions $P_1(\pi_1)$ and $P_2(\pi_2)$ when 
the number of plays $n$ is large and $n \simeq n_1 \gg n_2$. By the law of large numbers, the sample means $\hat{\pi}_i=w_i/n_i$ 
of the $\pi_i$'s converge to their respective values $p_i$
in the limit of large $n$. It follows that typically $\hat{\pi}_1 > \hat{\pi}_2$, as in Fig.~\ref{fig:infop}. The distribution $\rho(\pi_{\max})$ matches to a large extent the first term in the right hand side of \eqref{eq:rho} except at the right tail, where  the contribution by $\pi_2$ dominates as $n_2\ll n_1$. The right tail corresponds to the unlikely event that 
the inferior sample mean $\hat{\pi}_2$ is due to bad luck. Large deviations theory (see \cite{Cover}) ensures that the probability for $\hat\pi_2$ to be generated by a true probability of success $>p_1$, is exponentially small in $n_2$, as we discuss below. 

The differential entropy of the continuous distribution $\rho(\pi_{\max})$ is $H(\pi_{\max}) = -\int \rho(p) \ln \rho(p) dp$ -- we shall be interested in the increments of the entropy so that normalization (see Chap. 8 in \cite{Cover}) is not an issue here. 
The Info-p strategy chooses the arm which maximizes the expected reduction of entropy  $H$. Specifically, 
the expected reduction $\langle \Delta H \rangle_i$ upon playing the $i$th arm with the posterior $P_i$ given by \eqref{eq:posterior_beta} is\,: 
\begin{eqnarray}
&\langle \Delta H \rangle_i = \text{Pr}(\text{0 observed} |P_i) \times \Delta H(\pi_{\max}|\text{0 observed})+\nonumber \\ & \text{Pr}(\text{1 observed} | P_i) \times \Delta H(\pi_{\max}|\text{1 observed})\,,
\label{eq:DeltaH}
\end{eqnarray}
where $\text{Pr}(X\, \text{observed} | P_i) = \int P_i(p) \Pr(X|p_i = p)dp$ is the likelihood of $X=1/0$, which denote win/loss, respectively. The increments $\Delta H(\pi_{\max}|\text{X observed})$ are calculated by updating the posterior \eqref{eq:posterior_beta} appropriately, e.g. if $X=1$ then $n_i\mapsto n_i+1$ and $w_i\mapsto w_i+1$. The corresponding distribution $\rho(\pi_{\max})$ is then obtained using \eqref{eq:rho} and the increment of the entropy is finally calculated using the definition of $H(\pi_{\max})$  above.

The first arm of the bandit typically gives the dominant contribution to the entropy and Info-p plays it most frequently. However, as $n_1$ increases, the expected variation \eqref{eq:DeltaH} of the first arm diminishes and the second arm is eventually played, as we proceed to discuss analytically and numerically.

\subsection{Optimality of Info-p}  In the region around the sample mean $\hat{\pi}_1\simeq p_1$, the distribution $\rho(\pi_{\max})$ in \eqref{eq:rho} can be written as $\rho(\pi_{\max}) \simeq P_1(\pi_{\max}) \text{Pr}(\pi_2 < \pi_{\max}) \approx  P_1(\pi_{\max})$ where  $P_1$ is approximately normal due to the central limit theorem, and its variance $\sigma_1^2 = \hat{\pi}_1(1 - \hat{\pi}_1)/n_1$. 

The right tail of $\rho$ away from $\hat{\pi}_1$ is controlled by the theory of large deviations \cite{Cover}. Specifically, the probability that a sequence of outcomes with sample mean $\hat{\pi}_2$ is generated by a distribution with parameter $p$ is $e^{-n_2D(\hat{\pi}_2,p)}$, where the Kullback-Leibler divergence $D$ was defined above, see \eqref{eq:lairob}.  It follows from \eqref{eq:rho} that 
the right tail of $\rho(\pi_{\max}) \propto e^{-n_2D(\hat{\pi}_2, \pi_{\max})}\text{Pr}(\pi_1 < \pi_{\max}) \approx e^{-n_2D(\hat{\pi}_2, \pi_{\max})}$ where the second approximation holds for $\pi_{\max}> \hat{\pi}_1$ as the distribution of $\pi_1$ is strongly localized around its sample mean $ \hat{\pi}_1$. Ignoring subdominant terms, the contribution to the entropy $H(\pi_{\max})$ is 
$\propto \int_{\hat{\pi}_1}^1 n_2D(\hat{\pi}_2, p)e^{-n_2D(\hat{\pi}_2, p)}dp$. For moderately large $n_2$, the integral is dominated by the maximum of the exponential term and Laplace method gives
$\simeq n_2D(\hat{\pi}_2, \hat{\pi}_1)e^{-n_2D(\hat{\pi}_2, \hat{\pi}_1)}$. 

Adding up the two previous contributions, we conclude that
\begin{equation}
\label{eq:infopent}
H(\pi_{\max}) \approx \frac{1}{2}\ln 2\pi e \sigma_1^2 +  Ae^{-n_2D(\hat{\pi}_2, \hat{\pi}_1)}\,,
\end{equation}
%H(p_{\text{max}}) &\approx \underbrace{\frac{1}{2}\ln 2\pi e \sigma_1^2}_{\text{entropy of Gaussian body}} +  \underbrace{Ae^{-n_2D(\hat{p}_2, \hat{p}_1)}}_{\text{entropy of the tail}}
where $A$ is a subdominant prefactor. The first term on the right-hand side of \eqref{eq:infopent} becomes smaller as the first arm is played due to $\sigma_1^2\propto 1/n_1$, i.e. it is the exploitative term that selects the arm with the highest sample mean. The second, exploratory term in the right-hand side of \eqref{eq:infopent} 
accounts for the probability of misclassification, and it reduces as $n_2$ increases.  
%The form of the entropy in \eqref{eq:infopent} is not a coincidence; if we use the exact relation $H(\pi_{\max}) =H(\pi_{\max} | b_{\max}) + H(b_{\max}) - H(b_{\max}|\pi_{\max})$ and observe that in the asymptotic limit the density of $\pi_{\max}$ has distinct contributions from each bandit at a particular value of $\pi_{\max}$ i.e., $ H(b_{\max}|\pi_{\max})$ is zero, then $H(\pi_{\max}) \simeq qH(\pi_1) + (1-q)H(\pi_2) + H(b_{\max}) $ .

The neutral decision boundary, i.e. the boundary where the expected reduction of entropy on playing either arm is equal, is calculated for large $n$ by equating the variation of the two contributions in \eqref{eq:infopent}. By using 
$n_1 \simeq n$ and neglecting subdominant prefactors, we find
\begin{equation}
\ln n \simeq n_2D(\hat{\pi}_2, \hat{\pi}_1) + O(\ln n_2)\,.
\label{eq:decisionbound}
\end{equation}
In the limit of large $n$, the sample means tend to their respective values $p$'s and \eqref{eq:decisionbound} 
coincides with the Lai-Robbins bound \eqref{eq:lairob}. This establishes the optimality of Info-p, which we shall also verify numerically in the next Section.

Asymptotic optimality is intuited as follows. The order between sample means, say $\hat{\pi}_2<\hat{\pi}_1$, might be due to fluctuations and  we ought to make sure that the true probabilities of success are not inverted, i.e. that $p_2<p_1$. The probability of inversion is $\exp\left[-n_2D(\hat{\pi_2},\hat{\pi}_1)\right]$ by large-deviations theory \cite{Cover}. The 
exponential dependence on $n_2$ pushes toward $n_2\propto n$ whilst short-term reward pushes to play greedily the arm with the highest sample mean. The optimal trade-off is 
dictated by marginality of sampling\,: 
the number $n/n_2$ of possible stretches of size $n_2$ times the probability of inversion 
should satisfy $n/n_2\times \exp\left[-n_2D(\hat{\pi_2},\hat{\pi}_1)\right]\lesssim 1$. The dominant order of this expression yields the Lai-Robbins inequality \eqref{eq:lairob} and the marginal case defines the Info-p decision boundary \eqref{eq:decisionbound}.

\begin{figure*}
\begin{center}
\includegraphics[width=.8\textwidth]{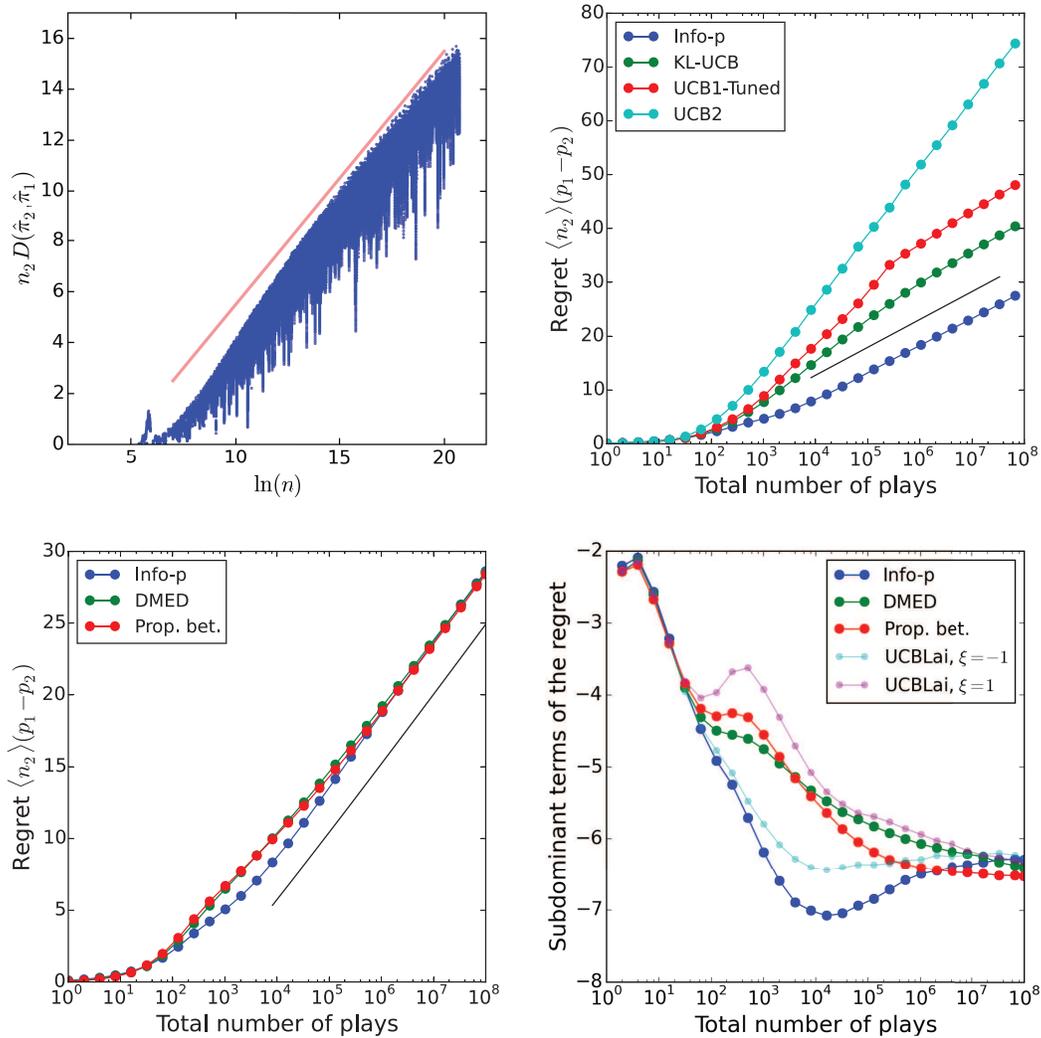}
\caption{Performance of Info-p and comparison with other strategies of decision. The two-armed bandit has $p_1=0.9$, $p_2=0.8$ as in Fig.~1.
The upper left panel shows points (blue dots) when Info-p played the second suboptimal arm. The plot  shows about 60,000 points cumulated over 250 realizations, each one of them lasting $n=10^9$ plays. The red line shows 
a line of slope one, which corresponds to the Info-p decision boundary \eqref{eq:decisionbound} in the asymptotic regime of large $n$. The upper right panel shows the comparison between the average regret obtained by Info-p and by the three Upper-Confidence Bound (UCB) strategies UCB-Tuned \cite{auer}, UCB2 \cite{auer}, KL-UCB \cite{cappe}.  The 
UCB strategies exhibit logarithmic but suboptimal regrets, which are manifestly asymptotically bigger as compared to Info-p. The lower panels show a comparison between the average regret obtained by Info-p and known asymptotically optimal decision strategies (DMED  \cite{honda}, Kelly's proportional betting \cite{Kelly} and UCB index policies defined in \cite{changlai}) discussed in the text.  Asymptotic optimality of all the strategies is visible in the left panel (the black line has the optimal slope $\ln n/D(p_2,p_1)$ 
in \eqref{eq:lairob}). In the right panel, we averaged over 25,000 statistical realizations and subtracted the dominant logarithmic term to evidence subdominant contributions\,: Info-p compares favorably with other algorithms and even features the smallest regret at intermediate times.} 
\label{fig:regret}
\end{center}
\end{figure*}

\subsection{Numerical simulations of Info-p}For our simulations, we chose a two-armed bandit with $p_1 = 0.9$, $p_2 = 0.8$. At every decision event, 
we compute the expected variation in entropy and choose the arm to play as described in \eqref{eq:DeltaH}. 
%\begin{align}
%\begin{split}
%\langle \Delta H \rangle_i = &\text{Pr}(\text{0 is observed} | \pi_i) \times \Delta H(\text{0 is observed})\\  &+ \text{Pr}(\text{1 is observed} | \pi_i) \times \Delta H(\text{1 is observed}) 
%\end{split}
%\end{align}
%where $\text{Pr}(o \text{ is observed} | \pi_i) = \int P_i(p) \Pr(o|p_i = p)dp$ is the expected likelihood of observation $o$. After receiving a sample from the chosen bandit that maximizes loss of entropy, the posterior of the bandit is updated using Bayes rule. 
Subdominant corrections to the regret are $O(\ln \ln n)$ \cite{lairob}. Consequently, clean data for the asymptotic regime require $n \gtrsim 10^6$, which is computationally demanding due to the updates of the posteriors at every step. 

To simulate the asymptotic regime, we developed an {\em exact} numerical technique (see Appendix \ref{a1}) that dramatically speeds up simulations. The logarithmic dependence in \eqref{eq:lairob} implies that asymptotically optimal strategies play long stretches of the estimated best arm, punctuated by short stretches of suboptimal arms. We derive then a rigorous lower bound for the duration of the long stretches and generate a single random variable for the cumulated reward over the entire stretch. 

Using the technique above, we verified that the decision boundary is indeed consistent with the optimality of Info-p\,: Fig.~\ref{fig:regret}A confirms that the points where Info-p chose the subdominant arm are below the predicted decision boundary  \eqref{eq:decisionbound} and approach it as $n$ increases. 

\subsection{Comparison between Info-p and other strategies of decision} The goal of this Section is to first briefly introduce state-of-the-art decision strategies whose regret increases logarithmically with $n$, and then compare them with Info-p.

Kelly's proportional betting \cite{Kelly} (known as Thompson sampling \cite{Kaufmann} in the machine learning community) is a randomized Bayesian strategy that plays arms with a probability proportional to their respective probability to be the best. Its asymptotic optimality was  recently proved in Ref.~\cite{Kaufmann} (see also Appendix \ref{a4}). 

Upper Confidence Bound (UCB) strategies are based on an index policy, like Gittins' index \eqref{eq:Gittins}, yet the calculation of the index is vastly simplified. Specifically, UCBs are formed by inflating the sample mean estimate of the probability of success of an arm with an additional positive term that accounts for the uncertainty in that estimate.  
A notable example is the UCB index $\chi_i$ introduced in Ref.~\cite{changlai}\,: if the $i$th arm was played $n_i$ times and its sample mean is $\hat{\pi}_i$, the index $\chi_i$ is defined via\,:  
$n_i D(\hat{\pi}_i,\chi_i) = \ln n/n_i + \xi \ln \ln n/n_i$, with $\chi_i > \hat{\pi}_i$. The constant $\xi$ 
generalizes the value $\xi=-1/2$ found by considering the Gittins' index \eqref{eq:Gittins} for Gaussian rewards in the limit $\gamma\to 1$  \cite{lai}. The class of models above is asymptotically optimal \cite{lai,changlai}. The value of $\xi$ is chosen empirically and controls subdominant terms. 

Figure~\ref{fig:regret}B shows the comparison between Info-p and UCB-Tuned \cite{auer}, UCB2 \cite{auer}, KL-UCB \cite{cappe}, which all 
exhibit logarithmic regret. However, their prefactor does not saturate the Lai-Robbins bound \eqref{eq:lairob} and UCB regrets are  asymptotically bigger as compared to Info-p.

Figures~\ref{fig:regret}C-D present a comparison of the regret {\it vs} $n$ for Info-p,  Kelly's proportional betting \cite{Kelly}, the UCB strategy DMED \cite{honda} and the UCBLai index policy \cite{changlai} for various values of its free parameter $\xi$. All the algorithms are asymptotically optimal and Info-p compares quite favorably with the others, especially at early and intermediate times when its regret remains below other curves. 

\subsection{Information about the identity of the best arm}

Infomax approaches can pursue information about diverse quantities. For multi-armed bandits, choosing which arm to play requires {\it a priori} only the identity of the best arm and not its probability of success. It is then natural to investigate the alternative Infomax approach that maximizes the information gain about the 
identity $b_{\text{max}}$ of the best arm. This possibility, which was previously mentioned in Ref.~\cite{Wyatt},  
is analyzed in detail in the next Section. Here, we determine the maximum
possible rate of information gain on $b_{\text{max}}$. 

The estimated probability for the  $i$-arm to be the best is denoted $q_i$. For two-armed bandits, $q_2+q_1=1$ and
\begin{equation}
q_2 = \int_0^1 P_1(p) dp \int_p^{1} P_2(q) dq\,.
\label{eq:q1}
\end{equation}
The posterior distributions $P_i$ are given by \eqref{eq:posterior_beta}. 

The entropy of the unknown identity $b_{\max}$ of the best arm is
$H(b_{\text{max}}) = -q_1\ln q_1 - q_2\ln q_2$. 
We are interested in the asymptotic limit of $n_1$ and $n_2$ large. Sample means $\hat{\pi}_i = \frac{w_i}{n_i} $ are then close to their true values $p_i$ and typically satisfy $\hat{\pi}_1 >  \hat{\pi}_2$. It follows that $q_1$ is close to unity and 
\begin{equation}
H(b_{\text{max}}) \sim -q_2\ln q_2\,.
\label{eq:Hbmax}
\end{equation}

The integrals that define $q_2$ in \eqref{eq:q1} have three contributions\,:

\noindent (I) The region $p\le\hat{\pi}_2$. There, we have $\int_p^{1} P_2(q) dq\sim 1$ and $P_1(p)\sim \exp\left[-n_1D\left(\hat{\pi}_1,p\right)\right]$ by large deviations theory \cite{Cover}. Integrating over $p$ and using that the dominant contribution comes from 
$p\simeq \hat{\pi}_2$, we obtain $\exp\left[-n_1D\left(\hat{\pi}_1,\hat{\pi}_2\right)\right]$.

\noindent (II) The region of $p$'s between $\hat{\pi}_2$ and $\hat{\pi}_1$. Its contribution is $\int \exp\left[-n_1D\left(\hat{\pi}_1,p\right)-n_2D\left(\hat{\pi}_2,p\right)\right]\,dp$ by large deviations theory. The integral can be calculated by Laplace method (see below) and we denote by $\pi_s$ the point where the maximum of the exponent is achieved. 

\noindent (III) Finally, the contribution from the rightmost region of $p$'s is dominated by $p\simeq \hat{\pi}_1$ and reads $\exp\left[-n_2D\left(\hat{\pi}_2,\hat{\pi}_1\right)\right]$.

\smallskip
In summary, we obtain
\begin{equation}
q_2\sim e^{-n_1D\left(\hat{\pi}_1,\hat{\pi}_2\right)}+e^{-n_1D\left(\hat{\pi}_1,\pi_s\right)-n_2D\left(\hat{\pi}_2,\pi_s\right)}+
e^{-n_2D\left(\hat{\pi}_2,\hat{\pi}_1\right)}\,,
\label{eq:q2asymp}
\end{equation}
where $\pi_s=\left(n_1\hat{\pi}_1+n_2\hat{\pi}_2\right)/n$ (see Appendix \ref{a2}). 

To minimize $\ln q_2$ -- thereby achieving the maximum acquisition of information, see \eqref{eq:Hbmax} -- we must  extremize with respect to $n_1$ and $n_2$. We show in Appendix \ref{a2} that the dominant contribution comes 
from the second exponential term in \eqref{eq:q2asymp}. The resulting extremum (with $n_1+n_2=n$) gives $D\left(\hat{\pi}_1,\pi_s\right)=D\left(\hat{\pi}_2,\pi_s\right)$. An important consequence of this equality is that $\pi_s$ is at a finite distance from $\hat{\pi}_1$ and $\hat{\pi}_2$. It follows then from the expression \eqref{eq:q2asymp} of $\pi_s$ that $n_2\propto n$, which is also confirmed by explicit expressions derived in the Appendix. 
As for the fastest rate of decay of the average logarithm of the entropy, we obtain\,:
\begin{equation}
\overline{\ln H(b_{\text{max}})}=-n D(p_1,p_s)=-nD(p_2,p_s)\,,
\label{eq:slopelogH}
\end{equation}
where $p_s$ is defined by the equality $D(p_1,p_s)=D(p_2,p_s)$.

\subsection{Info-id}
\begin{figure*}
\begin{center}
\includegraphics[width=.8\textwidth]{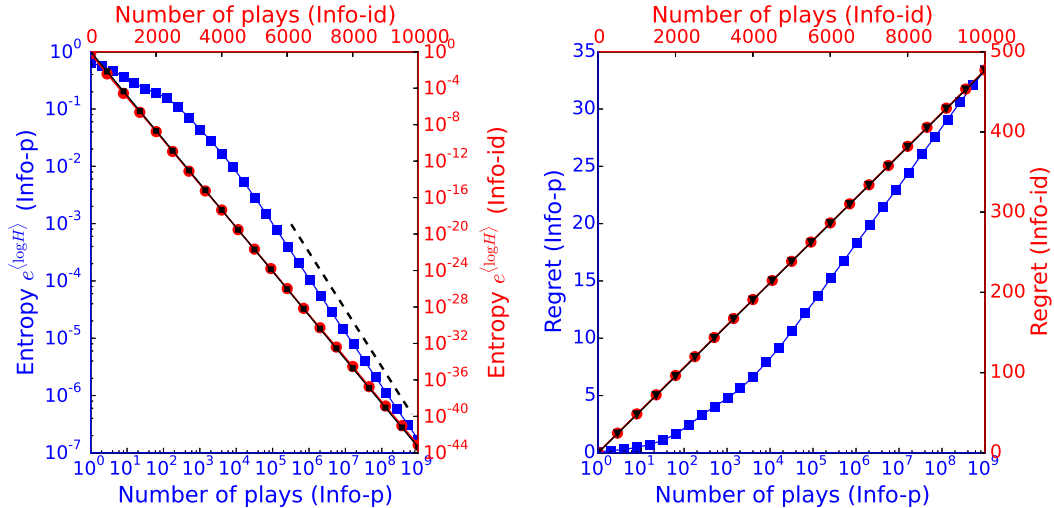}
\caption{The cost of information. A comparison of the entropy $H(b_{\max})$ on the identity of the best arm in the bandit (left panel) and the corresponding regret (right panel) for Info-id (red curves and axes) and Info-p (blue curves and axes). The left panel is in log-log scale for the blue curve and lin-log for the red curve. The right panel is in log-lin scale for the blue curve and lin-lin for the red curve. Info-id achieves the fastest possible reduction of the entropy, as shown by the agreement with the optimal slope \eqref{eq:slopelogH} (black solid line in the left panel). For Info-p, the decrease is much slower\,: $\propto 1/n$ (dashed  line). Conversely,  Info-id has a linear regret that largely exceeds the optimal Lai-Robbins bound \eqref{eq:lairob} achieved by Info-p.}
\label{fig:cost}
\end{center}
\end{figure*}

The Info-Id algorithm is defined as the decision strategy that chooses the arm which maximizes the expected reduction of $\ln H(b_{\text{max}})$. Specifically, the expected reduction $\langle \Delta \ln H\rangle_i$ upon 
playing the $i$-th arm with posterior $P_i$ is analogous to \eqref{eq:DeltaH} with $\Delta H(\pi_{\max})$ replaced by $\Delta \ln H(b_{\text{max}})$.
Increments are calculated by updating the posterior as for \eqref{eq:DeltaH}, by using \eqref{eq:q1} and finally obtaining the increment using the definition of $H(b_{\text{max}})$ above. This greedy, one-step-in-time procedure indeed achieves the fastest decrease \eqref{eq:slopelogH}
of $\overline{\ln H(b_{\text{max}})}$ (see Fig.~\ref{fig:cost}). Note the logarithm in the definition of Info-id\,: 
maximizing the expected reduction of $H(b_{\text{max}})$ would not achieve \eqref{eq:slopelogH} but a slower decay (see Appendix \ref{a2}).

\subsection{The cost and value of information} 
Information and payoffs embody the two sides of the exploration/exploitation dilemma for multi-armed bandits.
The expressions \eqref{eq:Hbmax}, \eqref{eq:q2asymp} and \eqref{eq:slopelogH} allow to quantify the trade-offs in the optimal behaviors achieved by Info-p and Info-id, respectively. 

The three relations above show that $-\overline{\ln H(b_{\text{max}})}\propto n\propto n_2$ for Info-id. Since 
regret is proportional to $n_2$, we conclude that the rate of decay $-\overline{\ln H(b_{\text{max}})}$ is proportional 
to the average regret, i.e. the exponential rate  \eqref{eq:slopelogH} implies a regret linear in $n$, as confirmed by numerical simulations (see Fig.~\ref{fig:cost}). 
%In words, the exponentially decaying entropy \eqref{eq:slopelogH} costs a largely suboptimal regret as compared to \eqref{eq:lairob}.  

A very different trade-off underlies the Info-p optimal regret. Indeed, for $n_2\propto \ln n$, the dominant contribution in \eqref{eq:q2asymp} comes from the last term and implies a power-law decay of the entropy. In particular, if the Lai-Robbins bound \eqref{eq:lairob} is saturated then $\overline{\ln H(b_{\text{max}})}\sim -\ln n$. 
The information on the identity of the best arm is therefore reducing much more slowly as compared to Info-id.  

\smallskip
The behaviors above clearly illustrate the costs in regret of  reducing $H(b_{\text{max}})$. 
However, information about $b_{\text{max}}$ has a definite value that can be exploited to 
increase payoffs. In particular, if we start playing with some {\it a priori} information on the identity of the best arm, i.e. $H(b_{\text{max}})=H_0<\ln 2$, general distortion-type arguments \cite{Cover,Bialek} suggest that payoffs should increase as $H_0$ reduces. 

\begin{figure*}
\begin{center}
\includegraphics[width=.8\textwidth]{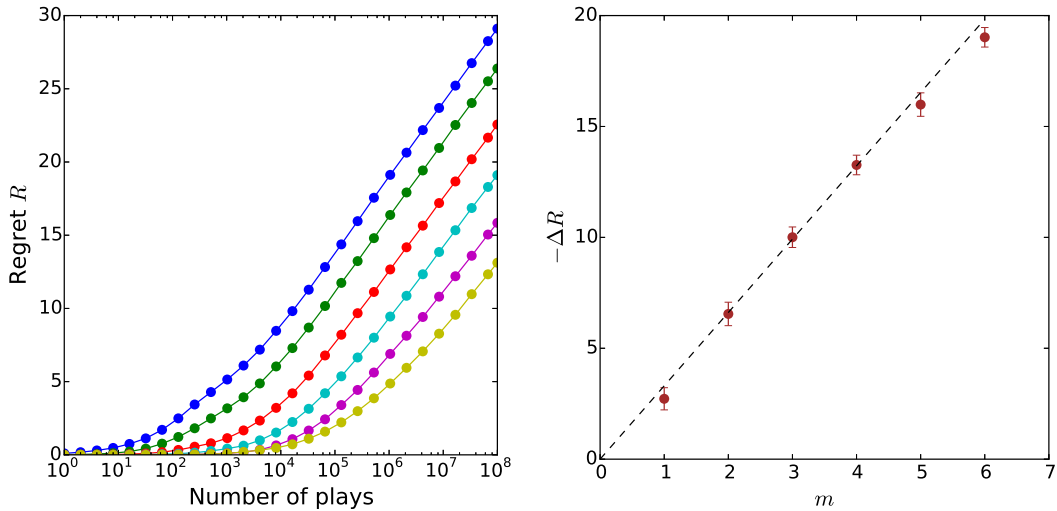}
\caption{The value of information. The left panel shows the curves for the regret obtained by Info-p after a period of ``pre-training'' by Info-id. The pre-training lasts for a number of iterations such that the initial entropy on the identity of the best arm $H(b_{\max}) = H_0$. Details are described in the text. From top to bottom (blue to yellow), the initial entropies are $H_0 = \ln 2/2^m$ for $m = 0,1,\dots,5$. The regret curves are parallel in the asymptotic regime, i.e. a multiplicative change in $H_0$ leads to a constant change in the regret. In the right 
panel we show that the corresponding rate-distortion curve for the reduction in regret $\Delta R$ measured at $n = 10^8$ for 
different values of the index $m$ in $H_0=\ln 2/2^m$. The approximate linearity of the rate-distortion curve 
$-\Delta R\propto m$ is captured (dashed line) by theoretical arguments presented in the main text and appendix.}
\label{fig:value}
\end{center}
\end{figure*}

We quantify the value of information by measuring the variation in payoff as a function of $H_0$. To generate the initial {\it a priori} information, we first play the bandit using Info-id until $H(b_{\text{max}})=H_0$ is achieved. Then, we switch to Info-p to compute the regret obtained with those pre-trained priors. Fig.~\ref{fig:value} confirms that information on $b_{\text{max}}$ has indeed a positive value,
and shows the rate-distortion curve for the variation $\Delta R$ of regret {\it vs} the initial entropy $H_0$. 

The rate-distortion curve in Fig.~\ref{fig:value} is rationalized as follows (see Appendix \ref{a3} for details). The Info-id ``pre-training'' required to reach $H(b_{\text{max}})=H_0=\ln 2/2^m$ lasts for $n^{(pt)}\propto m$ steps (see \eqref{eq:slopelogH}). Since $n_1$ and $n_2$ are both $\propto n$ for Info-id, the typical prior resulting from the pre-training is equivalent to the 
unlikely (for Info-p) situation of a comparable number of plays $n_1^{(pt)}$ and $n_2^{(pt)}$ for the two arms. Info-p will then play a very long stretch on the first arm until its typical decision boundary \eqref{eq:decisionbound} is reached. The corresponding reduction in average regret $-\Delta R$ is proportional to the logarithm of the length of the stretch, i.e. $n_2^{(pt)}\propto m$. We conclude that $-\Delta R\propto m$, as observed in Fig.~\ref{fig:value}.

\section{Conclusion}
We investigated the multi-armed bandit problem \cite{gittins} with the purpose of gaining insights 
on Infomax approaches, which postulate a functional role for the acquisition and transmission of information. We introduced two 
Infomax strategies of decision, and evaluated their performance using 
known results on optimal decisions for multi-armed bandits. 

The first strategy, Info-id, optimally acquires information on the identity of the best arm of the bandit but has a large, asymptotically linear regret. 
Note that the identity of the best arm is the quantity actually 
needed for the choice of the arm to play. Therefore, the first natural candidate for an Infomax approach is Info-id, which however 
performs poorly due to excessive exploration. 

The second strategy, Info-p, shifts the balance towards exploitation by gathering information on the highest expected reward among the arms. That pushes to play more frequently the estimated best arm in the bandit. We showed that this strategy yields asymptotically optimal regrets and compares favorably with state-of-the-art 
methods. The Info-p balance between exploration and exploitation produces a relatively slow $\propto 1/n$ acquisition of information on the identity of the best arm, which should be contrasted with the exponential decay achieved by Info-id. 
The striking differences between Info-p and Info-id clearly demonstrate that the nature of information 
acquired by Infomax critically matters. 

Info-p, like other Infomax approaches, uses information as a proxy, namely of cumulative payoffs for multi-armed bandits. As already mentioned in the Introduction, the advantage is that the proxy has general applicability and that the process of 
acquisition of information is one-step in time (and greedy in the choice among the options). 
The first point is important because situations where optimal policies are known are 
very rare. The second point is relevant because optimal policies often involve extended forecasts in the future, as shown by the example of the Gittins index \eqref{eq:Gittins}. Such calculations can {\it a priori} be formulated as dynamic programming yet in practice 
the size of the space to sample makes them unfeasible for computers and {\it a fortiori} for neural systems or single cells. 
It is therefore quite non-trivial that optimality on cumulative payoffs can be achieved by an Infomax approach for an appropriate quantity. That constitutes the general lesson 
drawn here\,: information in natural or artificial
systems could be acquired on quantities that are not 
immediately recognizable as functionally relevant but could actually allow for effective functional trade-offs. 

\begin{acknowledgements}
We are grateful to Boris Shraiman and Eric Siggia for illuminating discussions. 
MV acknowledges ICTP for hospitality and support. This work was supported by a grant from the Simons Foundation (\#340106, Massimo Vergassola). 
\end{acknowledgements} 

\appendix

\section{Fast Info-p numerical simulations}\label{a1}
Info-p is slowed down by posterior distributions \eqref{eq:posterior_beta} sharply peaking around their mean, and accuracy demands progressively finer discretization. To speed up the algorithm, we remark that the Lai-Robbins bound \eqref{eq:lairob} implies
that a typical play consists of long stretches of plays of the best arm interspaced with 
occasional plays of suboptimal ones. Suppose then the Info-p policy selects the best
 (with largest sample mean) arm for play. We can exactly bound the minimal length
of consecutive plays of the best arm as follows\,: Set the stretch size to some initial
guess\,; Consider the worst-case scenario of losses throughout the entire
stretch\,; If the Info-p policy chooses a sub-optimal arm at the end of the stretch, halve
the stretch size until the best arm is chosen\,; If the Info-p policy chooses the
best arm at the end of stretch, double the stretch size until a sub-optimal arm is chosen\,; Dissect dichotomically as in binary search algorithms \cite{Recipes} the 
intervals identified as above. Note that the worst-case
scenario of consecutive losses ensures that a lower bound on the stretch length is
obtained and the numerical technique is thereby exact.

Once the length of consecutive plays is identified, we generate a random variable for the number of wins during the stretch and update the posterior only once, by using the fact that $\beta$ distributions \eqref{eq:posterior_beta} are conjugate priors for 
Bernoulli likelihood functions \cite{MacKay}. Discretization of the state space is adaptive and refined as the number of plays increases so as to ensure proper accuracy. 
We employed a similar procedure for simulations of proportional betting (see \ref{a4}). 

\section{Proportional betting}\label{a4}
Kelly's proportional betting \cite{Kelly} (also called Thompson sampling in the machine learning community) is a randomized  policy that was recently shown to be asymptotically optimal \cite{Kaufmann}. At each step, the algorithm plays an arm with a probability proportional to its probability to be the best among the arms in the bandit. 

Our arguments for showing the optimality of Info-p (see main text) are easily adapted to confirm that proportional betting is indeed optimal. The probabilities for each arm to be 
the best are denoted $q_1, q_2,\ldots$. For two arms, in the asymptotic limit $n_1 \gg n_2$ and $n$ large, we typically have $\hat{\pi}_1 > \hat{\pi}_2$ and $\frac{n_1}{n_2} \approx \frac{q_1}{q_2}$ with $q_2 \simeq e^{-n_2D(\hat{\pi}_2, \hat{\pi}_1)}$ and $q_1 \simeq 1$, which again (as for Info-p) leads to $\ln n \simeq n_2D(\hat{\pi}_2, \hat{\pi}_1)$.  
 
Since proportional betting is a randomized algorithm, the technique used for the Fast Info-p algorithm (see Appendix \ref{a1}) does not carry over.  In the asymptotic limit, the probability that one of the arms is the best is very close to unity. This probability, say $q_1$, depends primarily on the number of plays of the inferior arms and changes negligibly as the first arm is played. Using this observation, the following approximate algorithm gives very reliable results: the best arm is played for a stretch whose size is randomly chosen from an exponential distribution of mean $\frac{1}{1-q_1}$. Immediately after the stretch, one of the inferior arms is chosen with probabilities $\frac{q_2}{1-q_1},\frac{q_3}{1-q_1},\dots$. This scheme is exact under the assumption that $q_1$ does not change during the stretch and we found it to be very reliable for the reasons mentioned above. The numerical method is analogous to the Gillespie algorithm  used to simulate chemical kinetics \cite{gillespie}. 

\section{Theoretical analysis of information on the identity of the best arm}\label{a2}
 The goal of this Section is to provide further details about optimal information on the identity of the best arm and the related Info-id policy. The policy greedily maximizes the reduction in log-entropy, $\ln H(b_{\max})$, where $H(b_{\max})$ is the entropy of the unknown identity $b_{\max}$ of the best arm in the bandit. As in the main text, we shall consider the case of a two-armed bandit with probabilities of success $p_1$ and $p_2$ ($p_1 > p_2$). Generalizations to bandits with more than two arms are straightforward.
 
The estimated values of the probabilities of success in a given sample of plays are denoted by $\pi_1$ and $\pi_2$, respectively. 
Their posterior distributions are given by \eqref{eq:posterior_beta}.
%\begin{equation}
%P_i(\pi_i) = \frac{\pi_i^{w_i} (1-\pi_i)^{n_i-w_i}}{B(w_i + 1, n_i- w_i + 1)}\,,
%\label{eq:posterior_betaSI}
%\end{equation}
%where $B$ denotes the Euler $\beta$-function. In \eqref{eq:posterior_betaSI} we have assumed a uniform prior\,; a different prior requires minor modifications and does not affect subsequent results. 
The sample mean of $\pi_i$ is indicated by $\hat{\pi}_i$. 

We denote by $q_1 = \text{Pr}(\pi_1 > \pi_2)$ the estimated probability for the first arm to be the best. For a two-armed bandit, $q_2 = 1-q_1$ and is given by \eqref{eq:q1}.
%\begin{equation}\label{eq:q1SI}
%q_2 = \int_0^1 P_1(p) dp \int_p^{1} P_2(q) dq\,,
%\end{equation}
%where the $P_i$'s are defined by \eqref{eq:posterior_betaSI}.  
The entropy of the unknown identity $b_{\max}$ of the best arm is\,: 
$H(b_{\text{max}}) = -q_1\ln q_1 -q_2\ln q_2$. In the asymptotic limit $n_1, n_2 \gg 1$, when the arms have each been played many times, the sample means $\hat{\pi}_i$ are typically close to their respective true values 
$p_i$. Large deviation theory \cite{Cover} states that the $i$th posterior and its cumulative distribution are both dominated by the exponential factor $e^{-{n_i}D(\hat{\pi}_i, p)}$. The probability $q_1$ is then close to unity and the entropy is well approximated by
\eqref{eq:Hbmax}. 
%\begin{align}
%H(b_{\max}) \simeq -q_2\ln q_2\,.
%\label{eq:H}
%\end{align}

When $n_1,n_2 \gg 1$, the integrals in \eqref{eq:q1} and \eqref{eq:Hbmax} can be calculated by Laplace method and have three contributions\,:

\noindent (I) The region $p\le\hat{\pi}_2$. There, we have $\int_p^{1} P_2(q) dq\sim 1$ and $P_1(p)\sim \exp\left[-n_1D\left(\hat{\pi}_1,p\right)\right]$ by large deviations theory \cite{Cover}. Integrating over $p$ and using that the dominant contribution comes from 
$p\simeq \hat{\pi}_2$, we obtain\,: $\exp\left[-n_1D\left(\hat{\pi}_1,\hat{\pi}_2\right)\right]$.

\noindent (II) The region of $p$'s between $\hat{\pi}_2$ and $\hat{\pi}_1$. Its contribution is $\int \exp\left[-n_1D\left(\hat{\pi}_1,p\right)-n_2D\left(\hat{\pi}_2,p\right)\right]\,dp$ by large deviations theory. Equating to zero the derivative of $n_1D(\hat{\pi}_1, p)+n_2D(\hat{\pi}_2, p)$ with respect to $p$ and using the definition of the Kullback-Leibler divergence $D(q,p) = q\ln \frac{q}{p} + (1-q)\ln \frac{1-q}{1-p}$, we obtain that the extremum is located at 
\begin{align}\label{eq:pisdef}
\pi_s = \frac{n_1\hat{\pi}_1 + n_2 \hat{\pi}_2}{n}\,,
\end{align}
where $n = n_1 + n_2$.

\noindent (III) Finally, the contribution from the rightmost region of $p$'s is dominated by $p\simeq \hat{\pi}_1$ and reads\,: $\exp\left[-n_2D\left(\hat{\pi}_2,\hat{\pi}_1\right)\right]$.

\smallskip
In summary, the asymptotic expression of the entropy is
\begin{align}\label{eq:entapprox}
H(b_{\max}) \sim A\exp\big[{-n_1D(\hat{\pi}_1, \hat{\pi}_2)}\big] + B\exp\big[{-n_2D(\hat{\pi}_2, \hat{\pi}_1)}\big] + C\exp\big[{-n_1D(\hat{\pi}_1, \pi_s)  -n_2D(\hat{\pi}_2, \pi_s)}\big]\,,
\end{align}
where $A,B,C$ are subdominant prefactors. 

\smallskip
The expression \eqref{eq:entapprox} still depends on $n_1$ and $n_2$, which are 
controlled by the policy of play. The fastest possible rate of acquisition of information is obtained by taking the extremum 
over $n_1$ and $n_2$ with the constraint $n_1+n_2=n$. Suppose for now (as we shall demonstrate  later) that the dominant contribution in \eqref{eq:entapprox} 
is the last one\,:
\begin{align}\label{eq:entdom}
H(b_{\max}) \sim \exp\big[{-n_1D(\hat{\pi}_1, \pi_s)  -n_2D(\hat{\pi}_2, \pi_s)}\big]\,.
\end{align}
The maximum possible rate of reduction of log-entropy is then calculated as follows. If we denote $n_1/n = x$, $n_2/n = 1-x$ and differentiate the exponent in \eqref{eq:entdom} with respect to $x$, we obtain the relation
\begin{align}\label{eq:optx}
D(\hat{\pi}_1, \pi_s) = D(\hat{\pi}_2, \pi_s)\,,
\end{align}
which defines the optimal value $\pi_s=\pi_{s,o}$. Using the explicit expression of the Kullback-Leibler divergence $D$\,:
\begin{equation}
\pi_{s,o} = \frac{1}{1 + e^{f(\hat{\pi}_1,\hat{\pi}_2)}}, \qquad f(\hat{\pi}_1,\hat{\pi}_2) = \frac{H(\hat{\pi}_1) - H(\hat{\pi}_2)}
{\hat{\pi}_1 - \hat{\pi}_2}\,.
 \label{eq:optpsi}
\end{equation}

The optimal proportion of plays on the arms follows from \eqref{eq:pisdef}\,:
\begin{align}
x_o=\left(\frac{n_1}{n}\right)_o = \frac{\pi_{s,o} - \hat{\pi}_2}{\hat{\pi}_1 - \hat{\pi}_2}\,. 
\label{eq:optn1}
\end{align}

The decay of the log-entropy averaged over the statistical realizations follows from \eqref{eq:entdom} and \eqref{eq:optx}\,:
\begin{equation}
\label{eq:optrate}
\overline{\ln H(b_{\rm max})} =-nD(p_1, p_{s,o})\,.
\end{equation}
where $p_{s,o} = (1 + e^{f(p_1,p_2)})^{-1}$ and $f$ is defined in \eqref{eq:optpsi}. Note that the average of the log-entropy gives the typical behavior over the realizations, while the entropy itself or its higher powers 
are determined by large-deviation fluctuations. That leads to anomalous exponents as a function of the power considered. The  appropriate statistic for the information gathered in a typical realization is $e^{\langle \ln H \rangle}$.

\medskip
The final piece of our analysis is to check that the claimed maximum exponent $-nD(\hat{\pi}_1, \pi_{s,o}) $ in \eqref{eq:entdom} 
is indeed larger than the other two potential candidates $-n x_oD(\hat{\pi}_1, \hat{\pi}_2)$ and $-n\left(1-x_o\right)D(\hat{\pi}_2, \hat{\pi}_1)$ in \eqref{eq:entapprox}\,:
\begin{align}
x_oD(\hat{\pi}_1, \hat{\pi}_2) \ge D(\hat{\pi}_1,\pi_{s,o})\,;\qquad (1-x_o)D(\hat{\pi}_2, \hat{\pi}_1) \ge D(\hat{\pi}_2,\pi_{s,o})\,.
\label{eq:ineq}
\end{align}
We concentrate on the first relation in \eqref{eq:ineq}\,; the second one follows by symmetry. The convexity of $D$ in the second argument implies\,:
\begin{align}
D(\hat{\pi}_1, \hat{\pi}_2) \ge D(\hat{\pi}_1, \pi_{s,o}) + (\hat{\pi}_2 - \pi_{s,o}) \times \frac{\pi_{s,o} - \hat{\pi}_1}{\pi_{s,o}(1-\pi_{s,o})}\,,
\label{eq:interm}
\end{align} 
where we used the explicit expression of the Kullback-Leibler divergence $D$ to calculate the partial 
derivative at $\pi_{s,o}$ with respect to the second argument. 
Multiplying by $x_o$ both sides of \eqref{eq:interm} and using \eqref{eq:optn1}, it follows that
\begin{align}
x_oD(\hat{\pi}_1, \hat{\pi}_2) \ge x_oD(\hat{\pi}_1, \pi_{s,o}) + \frac{\hat{\pi}_1 - \pi_{s,o}}{\hat{\pi}_1 - \hat{\pi}_2} \frac{(\hat{\pi}_2 - \pi_{s,o})^2}{\pi_{s,o}(1-\pi_{s,o})} \times \frac{D(\hat{\pi}_1,\pi_{s,o})}{D(\hat{\pi}_2,\pi_{s,o})}\,.
\end{align}
The ratio $\frac{D(\hat{\pi}_1,\pi_{s,o})}{D(\hat{\pi}_2,\pi_{s,o})} = 1$, due to \eqref{eq:optx}, and $\frac{\hat{\pi}_1 -\pi_{s,o}}{\hat{\pi}_1 - \hat{\pi}_2}=1-x_o$, due to \eqref{eq:optn1}. We conclude that\,:
\begin{align}
x_oD(\hat{\pi}_1, \hat{\pi}_2) \ge D(\hat{\pi}_1, \pi_{s,o}) + (1-x_o)D(\hat{\pi}_1,\pi_{s,o})\bigg[ \frac{(\hat{\pi}_2 - \pi_{s,o})^2}{\pi_{s,o}(1-\pi_{s,o})D(\hat{\pi}_2,\pi_{s,o})} - 1\bigg]\,.
\end{align}
To prove \eqref{eq:ineq}, it only remains to show that 
\begin{align}
\frac{(\hat{\pi}_2 - \pi_{s,o})^2}{\pi_{s,o}(1-\pi_{s,o})}\ge  D(\hat{\pi}_2,\pi_{s,o})\,,
\end{align} 
which follows from the inequality between the Kullback-Leibler  divergence and the $\chi^2$ distance of two distributions (see eqs.~6,7 in \cite{inequalities}). This completes the proof.

\subsection{A strategy that maximizes reduction in entropy}\,\,
Does the Info-id policy (which is greedy in its choice of the arm and one-step in time) attain the maximum rate \eqref{eq:optrate}\,? The aim of this subsection is to give a positive answer to this question.

\smallskip
The Info-d policy selects the arm of the bandit which offers the largest expected reduction in log-entropy
\begin{eqnarray}
&\langle \Delta \ln H \rangle_i = (1-\hat{\pi}_i) \times \Delta \ln H(b_{\max}|\text{0 observed})+\nonumber \\ & \hat{\pi}_i \times \Delta \ln H(b_{\max}|\text{1 observed})\,,
\label{eq:DeltalogH}
\end{eqnarray}
where $0/1$ correspond to loss/win and $\langle\bullet\rangle$ denotes the average with respect to the posterior probability distribution. To calculate $\langle \Delta \ln H \rangle_i$, we use the transformations: 
\begin{equation}\label{eq:transform}
 \text{0 is observed\,:}\left\{\begin{aligned}
        n_i&\rightarrow n_i + 1,\\
        \hat{\pi}_i&\rightarrow \hat{\pi}_i - \frac{\hat{\pi}_i}{n_i},\\
        \pi_s &\rightarrow \pi_s - \frac{\pi_s}{n}\,;
       \end{aligned}
       \right.
 \qquad 
 \text{1 is observed\,:}
  \left\{\begin{aligned}
        n_i&\rightarrow n_i + 1,\\
        \hat{\pi}_i&\rightarrow \hat{\pi}_i + \frac{1- \hat{\pi}_i}{n_i},\\
        \pi_s &\rightarrow \pi_s + \frac{1-\pi_s}{n}\,.
       \end{aligned}
 \right.
\end{equation}

Let us calculate the expected variation \eqref{eq:DeltalogH} upon playing the first arm,  $i=1$\,: 
\begin{eqnarray}
\label{eq:firstline}
& \ln H(b_{\max}|\text{0 observed}) \simeq -(n_1 + 1)D\bigg(\hat{\pi}_1 - \frac{\hat{\pi}_1}{n_1}, \pi_s - \frac{\pi_s}{n}\bigg) - n_2D\bigg(\hat{\pi}_2,\pi_s - \frac{\pi_s}{n}\bigg) \\ 
\label{eq:secondline}
&\simeq -(n_1+1)\bigg[D(\hat{\pi}_1 , \pi_s) - \frac{\hat{\pi}_1}{n_1}\ln \bigg( \frac{\hat{\pi}_1}{1- \hat{\pi}_1} \frac{1-\pi_s}{\pi_s}\bigg) - \frac{\pi_s}{n} \frac{\pi_s - \hat{\pi}_1}{\pi_s(1-\pi_s)}\bigg]
%& \hspace{2cm} 
- n_2\bigg[D(\hat{\pi}_2,\pi_s) - \frac{\pi_s}{n}\frac{\pi_s - \hat{\pi}_2}{\pi_s(1-\pi_s)}\bigg] \\
& \simeq -(n_1 + 1)D(\hat{\pi}_1 , \pi_s) - n_2D(\hat{\pi}_2,\pi_s) + \frac{n_1}{n}\frac{\pi_s - \hat{\pi}_1}{(1-\pi_s)} + \frac{n_2}{n}\frac{\pi_s - \hat{\pi}_2}{(1-\pi_s)} + \hat{\pi}_1\ln \bigg( \frac{\hat{\pi}_1}{1- \hat{\pi}_1} \frac{1-\pi_s}{\pi_s}\bigg) \label{eq:logH0_1}\,.
\label{eq:thirdline}
\end{eqnarray}
The first asymptotic equality \eqref{eq:firstline} follows from \eqref{eq:entdom} and \eqref{eq:transform}.
The second line \eqref{eq:secondline} is obtained by expanding $D(p,q)$ to first order in its Taylor series for both arguments, which  
is legitimate as $n_1,n_2\gg 1$. Finally, for the third line \eqref{eq:thirdline} we ignore subdominant terms $o(1)$. Notice that the sum of the third and the fourth terms in \eqref{eq:logH0_1} vanishes due to \eqref{eq:pisdef}. 

We conclude that 
\begin{equation}
\Delta \ln H(b_{\max}|\text{0 observed})\sim -D(\hat{\pi}_1 , \pi_s)  + \hat{\pi}_1\ln \bigg( \frac{\hat{\pi}_1}{1- \hat{\pi}_1} \frac{1-\pi_s}{\pi_s}\bigg)\,.
\label{eq:dlogH0}
\end{equation}
Similarly to \eqref{eq:dlogH0}, when the outcome of the play on the first arm is a win\,: 
\begin{eqnarray}
& \ln H(b_{\max}|\text{1 observed}) \sim -(n_1 + 1)D\bigg(\hat{\pi}_1 + \frac{1-\hat{\pi}_1}{n_1}, \pi_s + \frac{1-\pi_s}{n}\bigg) - n_2D\bigg(\hat{\pi}_2,\pi_s + \frac{1-\pi_s}{n}\bigg) \\ 
&\simeq -(n_1+1)D(\hat{\pi}_1 , \pi_s) - n_2D(\hat{\pi}_2,\pi_s) - \left(1-\hat{\pi}_1\right)\ln \bigg( \frac{\hat{\pi}_1}{1- \hat{\pi}_1} \frac{1-\pi_s}{\pi_s}\bigg)\,,
\label{eq:ssecondline}
\end{eqnarray}
where a cancellation similar to the one in \eqref{eq:thirdline} simplified the final expression \eqref{eq:ssecondline}. 
We are thereby left with
\begin{align}\label{eq:dlogH1}
\Delta \ln H(b_{\max}|\text{1 observed})\approx -D(\hat{\pi}_1 , \pi_s)  - (1-\hat{\pi}_1)\ln \bigg( \frac{\hat{\pi}_1}{1- \hat{\pi}_1} \frac{1-\pi_s}{\pi_s}\bigg)\,.
\end{align}
Finally, combining \eqref{eq:dlogH0} and \eqref{eq:dlogH1}, we obtain that
\begin{align}
\langle \Delta \ln H \rangle_1 = -D(\hat{\pi}_1 , \pi_s)\,. 
\end{align}
By symmetry, $\langle \Delta \ln H \rangle_2 = -D(\hat{\pi}_2 , \pi_s)$. We conclude that the decision boundary of Info-id matches the condition \eqref{eq:optx} and the policy indeed gathers information on the identity of the best arm at the maximum possible rate. 
 
\subsection{Why the variation of log-entropy rather than entropy?}

We stressed in the main text that Info-id is based on the expected variation of the log-entropy, as in \eqref{eq:DeltalogH}, and not the expected variation of the entropy. The reason is that the expected variation of 
the dominant term in \eqref{eq:entapprox} happens to vanish for the entropy. The choice of the arm to play is then based on subdominant terms, which yields a suboptimal rate as compared to \eqref{eq:optrate}.
The purpose of this subsection is to clarify this point.

Let us consider the expected variation of the entropy upon playing the $i$th arm\,: 
\begin{eqnarray}
&\langle \Delta H \rangle_i = (1-\hat{\pi}_i) \times \Delta H(b_{\max}|\text{0 observed})+ \hat{\pi}_i \times \Delta H(b_{\max}|\text{1 observed})\,,
\label{eq:DeltaHSI}
\end{eqnarray}
and consider first the third term \eqref{eq:entdom} (which is the one that gives the fastest possible 
decay \eqref{eq:optrate}). Using again the transformations \eqref{eq:transform}, its expected variation upon playing the first arm is
\begin{eqnarray}
\label{eq:boh}
&\langle \Delta \exp\big[{-n_1D(\hat{\pi}_1, \pi_s)  -n_2D(\hat{\pi}_2, \pi_s)}\big] \rangle_1 = (1-\hat{\pi}_1)\exp\big[-(n_1+1)D\big(\hat{\pi}_1 - \frac{\hat{\pi}_1}{n_1}, \pi_s - \frac{\pi_s}{n}\big) -  n_2D\big(\hat{\pi}_2, \pi_s - \frac{\pi_s}{n}\big)\big] \nonumber  \\ & \!\!\!\!\!\!+ \hat{\pi}_1\!\exp\big[\!-(n_1+1)D\big(\hat{\pi}_1\! +\! \frac{1-\hat{\pi}_1}{n_1}, \pi_s + \frac{1-\pi_s}{n} \big)\! -  n_2D\big(\hat{\pi}_2, \pi_s \!+\! \frac{1-\pi_s}{n}\big)\big]\! -\! \exp\left[-n_1D\left(\hat{\pi}_1, \pi_s\right) \!-\!  n_2D\left(\hat{\pi}_2, \pi_s\right)\right]\,.
\end{eqnarray}
Note that the exponents in the first two terms on the right-hand side of \eqref{eq:boh} are related to the objects that we calculated in the previous subsection. Using \eqref{eq:dlogH0} and \eqref{eq:dlogH1}, it follows then from \eqref{eq:boh} that 
\begin{eqnarray}
&& \langle \Delta \exp\big[{-n_1D(\hat{\pi}_1, \pi_s)  -n_2D(\hat{\pi}_2, \pi_s)}\big] \rangle_1 \propto \left\{(1-\hat{\pi}_1)\exp\bigg[-D(\hat{\pi}_1 , \pi_s)  + \hat{\pi}_1\ln \bigg( \frac{\hat{\pi}_1}{1- \hat{\pi}_1} \frac{1-\pi_s}{\pi_s}\bigg)\bigg]\right. \nonumber \\ 
 &
& \left. +\hat{\pi}_1\exp\bigg[ -D(\hat{\pi}_1 , \pi_s)  - (1-\hat{\pi}_1)\ln \bigg( \frac{\hat{\pi}_1}{1- \hat{\pi}_1} \frac{1-\pi_s}{\pi_s}\bigg) \bigg] - 1\right\} \,. \label{eq:dH_dom}
\end{eqnarray}
If the two terms \eqref{eq:dlogH0} and \eqref{eq:dlogH1} at the exponent in \eqref{eq:dH_dom} were small, then one would Taylor expand the exponentials and conclude that the variation of the entropy and the log-entropy are proportional. However, that is not the case because \eqref{eq:dlogH0} and \eqref{eq:dlogH1} are $O(1)$. By 
inserting the explicit form of the Kulback-Leibler divergence $D(p,q) = p\ln \frac{p}{q} + (1-p) \ln \frac{1-p}{1-q}$, the first and second terms on the right-hand side of \eqref{eq:dH_dom} actually reduce to $1-\pi_s$ and $\pi_s$, respectively. Therefore, the expected variation in the dominant term of the entropy turns out to vanish. 

\medskip
To determine the policy determined by the maximization of the expected decrease of entropy, we need then to consider  
subdominant terms in  \eqref{eq:entapprox}. Let us start with the first one\,:
\begin{align}
\langle \Delta \exp\big[-n_1D(\hat{\pi}_1, \hat{\pi}_2)\big] \rangle_1 &= (1-\hat{\pi}_1)\exp\big[-(n_1+1)D\big(\hat{\pi}_1 - \frac{\hat{\pi}_1}{n_1}, \hat{\pi}_2 \big)\big]  \nonumber  \\ & + \hat{\pi}_1\exp\big[-(n_1+1)D\big(\hat{\pi}_1 + \frac{1-\hat{\pi}_1}{n_1}, \hat{\pi}_2 \big) \big] -  \exp\big[-n_1D(\hat{\pi}_1, \hat{\pi}_2)\big]\,.  
\label{eq:bboh}
\end{align}
By Taylor expanding the Kullback-Leibler divergence as we have done previously, one can check that the right-hand side 
in \eqref{eq:bboh} is proportional to the right-hand side in \eqref{eq:dH_dom} and the expected variation for this term vanishes as well. 

The only non-vanishing contribution upon playing the first arm stems from the second term in \eqref{eq:entapprox}\,:
\begin{align}
\langle \Delta \exp\big[-n_2D(\hat{\pi}_2, \hat{\pi}_1)\big] \rangle_1 &= (1-\hat{\pi}_1)\exp\big[-n_2D\big(\hat{\pi}_2, \hat{\pi}_1 - \frac{\hat{\pi}_1}{n_1} \big)\big]  \nonumber  \\ & + \hat{\pi}_1\exp\big[-n_2D\big(\hat{\pi}_2, \hat{\pi}_1 + \frac{1-\hat{\pi}_1}{n_1}\big) \big] -  \exp\big[-n_2D(\hat{\pi}_2, \hat{\pi}_1)\big]\,.
\end{align}
Expanding again to first order in Taylor series, we get
\begin{align}
\langle \Delta \exp\big[-n_2D(\hat{\pi}_2, \hat{\pi}_1)\big] \rangle_1 &= \exp\big[-n_2D(\hat{\pi}_2, \hat{\pi}_1)\big] \bigg\{(1-\hat{\pi}_1)\exp\bigg[ \frac{\hat{\pi}_1n_2}{n_1} \frac{\hat{\pi}_1 - \hat{\pi}_2}{\hat{\pi}_1(1-\hat{\pi}_1)} \bigg] \\ & + \hat{\pi}_1 \exp\bigg[ -\frac{(1-\hat{\pi}_1)n_2}{n_1} \frac{\hat{\pi}_1 - \hat{\pi}_2}{\hat{\pi}_1(1-\hat{\pi}_1)} \bigg] \bigg\}\,.
\end{align}
The terms in the curly braces have $n_2$ and $n_1$ only as ratios and tend to {\em non-vanishing} 
constants in the asymptotic limit. 
The asymptotic behavior is therefore dominated by the exponential decay in $n_2$. The expected variation upon playing 
the second arm of the bandit is obtained by interchanging indices. We conclude that
\begin{align}
\langle \Delta H \rangle_1 \sim \exp\big[-n_2D(\hat{\pi}_2, \hat{\pi}_1)\big] \,;\qquad
\langle \Delta H \rangle_2 \sim \exp\big[-n_1D(\hat{\pi}_1, \hat{\pi}_2)\big]\,.
\label{eq:bbboh}
\end{align}

It follows from \eqref{eq:bbboh} that the behavior of the policy based on the maximization of the expected reduction of entropy 
depends on the balance between subdominant terms and that the decision boundary satisfies the relation 
\begin{align}\label{eq:dH_decisionbound}
n_1D(\hat{\pi}_1, \hat{\pi}_2) = n_2D(\hat{\pi}_2, \hat{\pi}_1)\quad \Rightarrow \quad \tilde{x}=\frac{D(\hat{\pi}_1, \hat{\pi}_2)}{D(\hat{\pi}_1, \hat{\pi}_2)+D(\hat{\pi}_2, \hat{\pi}_1) }\,.
\end{align}
The relations \eqref{eq:dH_decisionbound} should be contrasted with \eqref{eq:optx} and \eqref{eq:optn1}.

It remains to show that the decay of the average log-entropy generated by the policy \eqref{eq:dH_decisionbound} is still given by the third term in \eqref{eq:entapprox} with the exponent evaluated at $x=\tilde{x}$ (and not $x_o$ as for the optimal policy \eqref{eq:optx}). The inequality to be proved is\,:
\begin{align}
\tilde{x}D(\hat{\pi}_1, \tilde{\pi}_s) + (1-\tilde{x})D(\hat{\pi}_2, \tilde{\pi}_s) \le \tilde{x}D(\hat{\pi}_1, \hat{\pi}_2)= (1-\tilde{x})D\left(\hat{\pi}_2, \hat{\pi}_1\right)\,,
\end{align}
with $\tilde{\pi}_s=\tilde{x}\hat{\pi}_1+\left(1-\tilde{x}\right)\hat{\pi}_2$. The convexity in the second argument of the Kullback-Leibler 
divergence gives
\begin{align}
\tilde{x}D(\hat{\pi}_1, \tilde{\pi}_s) &\le \tilde{x}(1-\tilde{x})D(\hat{\pi}_1, \hat{\pi}_2), \\ (1-\tilde{x})D(\hat{\pi}_2,\tilde{\pi}_s) &\le (1-\tilde{x})\tilde{x}D(\hat{\pi}_2, \hat{\pi}_1)\,.
\end{align}
Summing up the two inequalities above and using \eqref{eq:dH_decisionbound}, the required relation is obtained. 

\begin{figure*}
\begin{center}
\includegraphics[width=.8\textwidth]{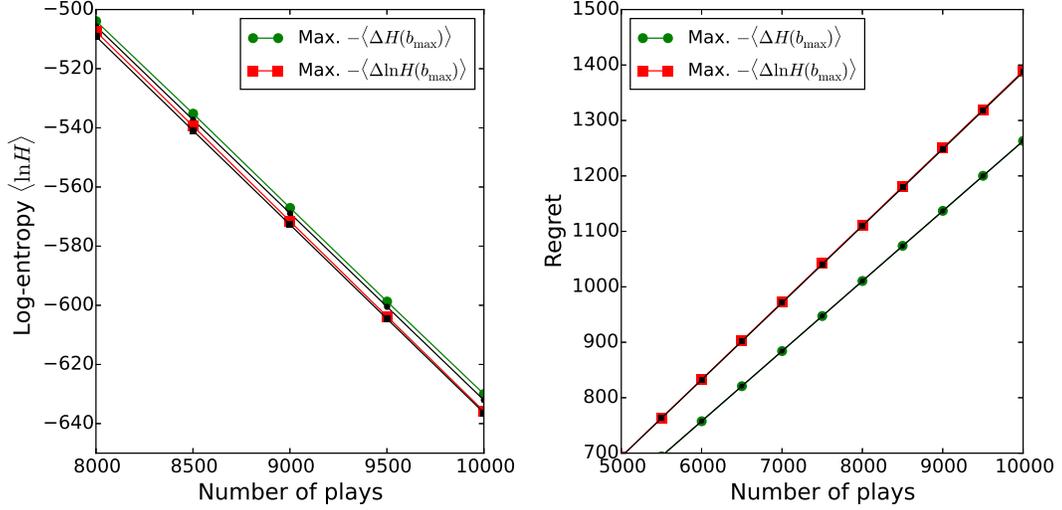}
\caption{The average log-entropy $\overline{\ln H(b_{\max})}$ and the average regret $R = \overline{n}_2(p_1-p_2)$ 
for the two policies that greedily maximize the expected reduction of the entropy ($-\langle\Delta H(b_{\max}) \rangle$) and 
the expected reduction of the log-entropy ($-\langle\Delta \ln H(b_{\max}) \rangle$). Numerical results from 
simulations are shown by green circles and red squares, respectively. Black lines with circular and square symbols are the corresponding theoretical predictions \eqref{eq:nonso} and \eqref{eq:optrate}, respectively. The values of the two probabilities of success are $p_1 = 0.9$ and $p_2 = 0.6$, differing from the ones used in the other figures in order to enhance the difference in entropies of the two strategies. The entropy for the Info-id strategy decays faster, although the difference is small. The regret of Info-id is bigger, as expected from the proportionality between average regret and rate of decay of $\ln H(b_{\max})$ discussed in the main text.}\label{fig:comparison}
\end{center}
\end{figure*}

In summary, the policy that maximizes the reduction of entropy (rather than the reduction of log-entropy) yields
\begin{equation}
\label{eq:nonso}
\overline{\ln H\left(b_{max}\right)} = -\frac{D\left(p_1,p_2\right)D(p_1,\tilde{p}_s)+D(p_2,p_1)D(p_2,\tilde{p}_s)}{D(p_1, p_2)+D(p_2,p_1) }\,;\qquad \tilde{p}_s=\frac{p_1D\left(p_1,p_2\right)+p_2D(p_2,p_1)}{D(p_1, p_2)+D(p_2,p_1)}\,.
\end{equation}
The decay is slower than for 
the optimal value  \eqref{eq:optx}, which was derived by extremizing over $x$ to obtain the optimal value $x_o$.
In Figure \ref{fig:comparison}, we confirm the theoretical predictions and compare the regret and the entropy for 
the two algorithms.  

\section{Quantifying the value of information}\label{a3}
The value of information is the reduction in the average regret obtained when some {\it a priori} information is available. In this section, we provide details on the theoretical argument sketched in the main text. The initial entropy of the identity of the best arm is supposed to be $H(b_{\max}) = H_0=\frac{\ln 2}{2^m}$. 

As mentioned in the main text, the ``pre-training'' with Info-id lasts for $n^{(pt)}$ steps. Since \eqref{eq:entdom} implies that 
$\ln H(b_{\max})=-nD(\hat{\pi}_1, \pi_{s,o})$ , the number of steps $n^{(pt)}$ satisfies $n^{(pt)}\simeq m\ln 2/D(\hat{\pi}_1, \pi_{s,o})$ with 
$\pi_{s,o}$ given by \eqref{eq:optpsi}. 

During the pre-training, the two arms are played $n_1^{(pt)}$  and $n_2^{(pt)}$ times. Their respective proportions are 
controlled by the expression \eqref{eq:optn1}. 
In particular, $n_2^{(pt)}=n^{(pt)}(\hat{\pi}_1-\pi_{s,o})/(\hat{\pi}_1-\hat{\pi}_2)$. Note that $n_2^{(pt)}$ scales linearly 
with $n^{(pt)}$ and is therefore much bigger than for typical Info-p statistics, where it would scale logarithmically with $n^{(pt)}$.

Since the suboptimal arm has been vastly overplayed in comparison with the typical Info-p statistics, once the algorithm switches to  Info-p after the pre-training, a long stretch of plays of the best arm will ensue. The length $\ell$ of the stretch is estimated by calculating the time taken to reach the Info-p decision boundary, i.e. $\ln \ell\sim n_2^{(pt)}D(\hat{\pi}_2,\hat{\pi}_1)$. 
In the absence of any pre-training, 
a stretch of length $\ell$ would lead to an average regret $R=(p_1-p_2)\ln \ell/D(p_2,p_1)$ (see the Lai-Robbins bound \eqref{eq:lairob} in the main text). We conclude that 
the expected difference in regret $\Delta R$ between the case with prior information and the case without, is given by 
%$\Delta R=-(p_1-p_2)\overline{n_2^{(pt)}}$. Combining the various expressions, we conclude that 
%We'll get the exact relation. From eq.\ref{eq:entdom}, the entropy $H_0$ after $n^{(pt)}$ plays is 
%\begin{align}
%H_0 &\propto e^{-n^{(pt)}D(\hat{\pi}_1, \pi_s)} \quad \text{i.e.,} \\ 
%-\ln H_0 &= n_2^{(pt)}\frac{D(\hat{\pi}_1, \pi_s)}{1-x^{(pt)}} - \ln\ln 2
%\end{align}
%where $x^{(pt)} = n_1^{(pt)}/n^{(pt)}$. The constant is as written above since when there is no pre-training, $n_2^{(pt)} = 0, H_0= \ln 2$. Using eq.\ref{eq:optn1}, we get
\begin{align}
\Delta R \simeq-\left(p_1-p_2\right)\overline{n_2^{(pt)}}\simeq -\ln 2\frac{p_1 -p_s}{D(p_1,p_s)}m\,,
\end{align}
with $p_s=\left(1+e^{f(p_1,p_2)}\right)^{-1}$ and the function $f$ defined by \eqref{eq:optn1}. The agreement with numerical simulations is shown in Fig.~4. Small deviations are ascribed to finite-size effects, e.g. the Info-p decision boundary that we used to determine the length $\ell$ of the initial stretch is only asymptotically valid, as evidenced in Fig.~2 (upper left panel). 

\bibliography{prx_bandits}   % name your BibTeX data base

\providecommand{\noopsort}[1]{}\providecommand{\singleletter}[1]{#1}%
\begin{thebibliography}{10}

\bibitem{Shannon48}
C.~E. Shannon.
\newblock The mathematical theory of communication.
\newblock {\em Bell Sys Tech J,}, 27:379--423, 1948.

\bibitem{Gallager}
R.~G. Gallager.
\newblock {\em Information Theory and Reliable Communication}.
\newblock Wiley, New York, 1968.

\bibitem{Mezard}
M.~M\'ezard and A.~Montanari.
\newblock {\em Information, Physics and Computation}.
\newblock Oxford University Press, Oxford, 2009.

\bibitem{MacKay}
D.~J.~C. MacKay.
\newblock {\em Information Theory, Inference and Learning Algorithms}.
\newblock Cambridge University Press, 2003.

\bibitem{Kelly}
J.~L. Kelly.
\newblock A new interpretation of information rate.
\newblock {\em Bell System Technical Journal}, 35:917--926, 1956.

\bibitem{Howard}
R.~A. Howard.
\newblock Information value theory.
\newblock {\em IEEE Trans Systems Science and Cybernetics}, 2:22--26, 1966.

\bibitem{Barron}
A.~Barron and T.~M. Cover.
\newblock A bound on the financial value of information.
\newblock {\em IEEE Trans Inf Theory}, 34:1097--1100, 1988.

\bibitem{Cover}
T.~M. Cover and J.~A. Thomas.
\newblock {\em Elements of Information Theory}.
\newblock Wiley, New York, second edition, 2006.

\bibitem{Bergstrom}
C.~T. Bergstrom and M.~Lachmann.
\newblock Shannon information and biological fitness.
\newblock {\em Proceedings of the IEEE Workshop on Information Theory}, 2004.

\bibitem{Bialek}
W.~Bialek.
\newblock {\em Biophysics: Searching for Principles}.
\newblock Princeton University Press, 2012.

\bibitem{Kussell}
E.~Kussell and S.~Leibler.
\newblock Phenotypic diversity, population growth, and information in
  fluctuating environments.
\newblock {\em Science}, 309:2075--2078, 2005.

\bibitem{Donaldson-Matasci}
M.~C. Donaldson-Matasci, C.~T. Bergstrom, and M.~Lachmann.
\newblock The fitness value of information.
\newblock {\em Oikos}, 119:219--230, 2010.

\bibitem{Rivoire}
O.~Rivoire and S.~Leibler.
\newblock The value of information for populations in varying environments.
\newblock {\em J Stat Phys}, 142:1124--1166, 2011.

\bibitem{Barlow61}
H.~B. Barlow.
\newblock {\em Possible principles underlying the transformation of sensory
  messages}, chapter~13.
\newblock MIT Press, 1961.

\bibitem{Laughlin89}
S.~B. Laughlin.
\newblock The role of sensory adaptation in the retina.
\newblock {\em J. Exp. Biol}, 146:39--62, 1989.

\bibitem{Atick92}
J.~J. Atick and A.~N. Redlich.
\newblock What does the retina know about natural scenes?
\newblock {\em Neural Computation}, 4:196--210, 1992.

\bibitem{Rieke}
F.~Rieke, D.~Warland, R.~Stevenick, and W.~Bialek.
\newblock {\em Spikes: Exploring the Neural Code}.
\newblock Bradford Book, 1999.

\bibitem{Dayan}
P.~Dayan and L.~F. Abbott.
\newblock {\em Theoretical Neuroscience: Computational and Mathematical
  Modeling of Neural Systems}.
\newblock MIT Press, Cambridge, 2001.

\bibitem{Cheong}
R.~Cheong, A.~Rhee, C.~J. Wang, I.~Nemenman, and A.~Levchenko.
\newblock Information transduction capacity of noisy biochemical signaling
  networks.
\newblock {\em Science}, 334:354--358, 2011.

\bibitem{Margolin}
A.~A. Margolin, I.~Nemenman, K.~Basso, C.~Wiggins, G.~Stolovitzky, and R.~D.
  Favera.
\newblock Aracne: An algorithm for the reconstruction of gene regulatory
  networks in a mammalian cellular context.
\newblock {\em BMC Bioinformatics}, 7, 2006.
\newblock Supp 1.

\bibitem{Francois}
P.~Fran\c{c}ois and E.~D. Siggia.
\newblock Predicting embryonic patterning using mutual entropy fitness and in
  silico evolution.
\newblock {\em Development}, 137:2385--2395, 2010.

\bibitem{Nemenman}
I.~Nemenman.
\newblock {\em Information theory and adaptation}.
\newblock Chapman and Hall/CRC Mathematical and Computational Biology. CRC
  Press, 2012.

\bibitem{Sharpee}
T.~O. Sharpee, A.~J. Calhoun, and S.~H. Chalasani.
\newblock Information theory of adaptation in neurons, behavior, and mood.
\newblock {\em Curr Opin Neurobiol}, 25:47--53, 2014.

\bibitem{Tkacik}
G.~Tkacik, C.~G.~Callan Jr., and W.~Bialek.
\newblock Information flow and optimization in transcriptional control.
\newblock {\em Proc. Natl. Acad. Sci. USA}, 105:12265--70, 2008.

\bibitem{Walczak}
G.~Tkacik and A.~M. Walczak.
\newblock Information transmission in genetic regulatory networks: a review.
\newblock {\em J. Phys.: Condens. Matter}, 23(15), 2011.

\bibitem{Linsker88}
R.~Linsker.
\newblock Self-organization in a perceptual network.
\newblock {\em IEEE Computer}, 21(3):105--117, 1988.

\bibitem{Bell95}
A.~J. Bell and T.~J. Sejnowski.
\newblock An information-maximization approach to blind separation and blind
  deconvolution.
\newblock {\em Neural Computation}, 7:1129--1159, 1995.

\bibitem{infotaxis}
M.~Vergassola, E.~Villermaux, and B.~Shraiman.
\newblock Infotaxis as a strategy for searching without gradients.
\newblock {\em Nature}, 445:406--409, 2007.

\bibitem{Tishby}
N.~Tishby and D.~Polani.
\newblock {\em Information theory of decisions and actions}, pages 601--636.
\newblock Springer, New York, 2011.

\bibitem{Sutton}
R.~Sutton and A.~Barto.
\newblock {\em Reinforcement Learning: An Introduction}.
\newblock MIT Press, Cambridge, 1998.

\bibitem{berry}
D.~A. Berry and B.~Fristedt.
\newblock {\em Bandit problems: sequential allocation of experiments}.
\newblock Springer, Dordrecht, 2001.

\bibitem{gittins}
J.~Gittins, K.~Glazebrook, and R.~Weber.
\newblock {\em Multi-armed Bandit Allocation Indices}.
\newblock John Wiley and Sons, second edition, 2011.

\bibitem{whittle}
P.~Whittle.
\newblock {\em Optimization over time, dynamic programming and stochastic
  control}.
\newblock Wiley Series in Probability and Statistics. 1982.

\bibitem{Gittins_paper}
J.~C. Gittins.
\newblock Bandit processes and dynamic allocation indices.
\newblock {\em Journal of the Royal Stat. Soc. B}, 6:148--177, 1995.

\bibitem{lairob}
T.~L. Lai and H.~Robbins.
\newblock Asymptotically efficient adaptive allocation rules.
\newblock {\em Advances in Applied Mathematics}, 6:4--22, 1985.

\bibitem{Burnetas97}
A.~N. Burnetas and M.~N. Katehakis.
\newblock Optimal adaptive policies for markov decision processes.
\newblock {\em Mathematics of Operations Research}, 22(1), 1997.

\bibitem{auer}
P.~Auer, N.~Cesa-Bianchi, and P.~Fischer.
\newblock Finite-time analysis of the multi-armed bandit problem.
\newblock {\em Machine Learning Journal}, 47:235--256, 2002.

\bibitem{cappe}
O.~Capp\'{e}, A.~Garivier, O.~Maillard, R.~Munos, and G.~Stoltz.
\newblock Kullback-leibler upper confidence bounds for optimal sequential
  allocation.
\newblock {\em Annals of Statistics}, 41(3):1516--1541, 2013.

\bibitem{honda}
J.~Honda and A.~Takemura.
\newblock An asymptotically optimal bandit algorithm for bounded support
  models.
\newblock {\em Proceedings of the Annual Conference on Learning Theory (COLT)},
  2010.

\bibitem{Kaufmann}
E.~Kaufmann, N.~Korda, and R.~Munos.
\newblock {\em Thompson Sampling: An Asymptotically Optimal Finite Time
  Analysis}, volume 7568 of {\em Lecture Notes in Computer Science}, pages
  199--213.
\newblock Springer, Berlin Heidelberg, 2012.

\bibitem{Wyatt}
J.~Wyatt.
\newblock {\em Exploration and Inference in Learning form Reinforcement}.
\newblock {Ph.D.} thesis, University of Edinburgh, 1997.

\bibitem{Recipes}
W.~H. Press, S.~A. Teukolsky, W.~T. Vettering, and B.~P. Flannery.
\newblock {\em The Art of Scientific Computing Numerical Recipes in C}.
\newblock Cambridge University Press, second edition, 1992.

\bibitem{changlai}
F.~Chang and T.~L. Lai.
\newblock Optimal stopping and dynamic allocation.
\newblock {\em Advances in Applied Probability}, 19(4):829--853, 1987.

\bibitem{lai}
T.~L. Lai.
\newblock Adaptive treatment allocation and the multi-armed bandit problem.
\newblock {\em Annals of Statistics}, 15(3):1091--1114, 1987.

\bibitem{inequalities}
T.~van Erven and O.~Harremoes.
\newblock R\'{e}nyi divergence and kullback-leibler divergence.
\newblock {\em IEEE Transactions on Information Theory}, 60(7):3797--3820,
  2014.

\bibitem{gillespie}
D.~T. Gillespie.
\newblock Exact stochastic simulation of coupled chemical reactions.
\newblock {\em Journal of Physical Chemistry}, 81(25), 1977.

\end{thebibliography}

%\begin{thebibliography}{}
% etc
%\end{thebibliography}

\end{document}